\documentclass[a4paper,runningheads]{llncs}
\usepackage[utf8]{inputenc}
\usepackage{graphicx}
\usepackage{xcolor}
\usepackage{soul}
\usepackage{booktabs}
\usepackage{amsmath}
\usepackage{amssymb}
\usepackage{mathtools}

\usepackage{hyperref}

\newif\ifrevision
\revisionfalse
\revisiontrue

\begin{document}

\title{A Straightforward Framework For Video Retrieval Using CLIP}

\author{
    Jes\'{u}s Andr\'{e}s Portillo-Quintero\orcidID{0000-0002-9856-1900} \and \\
    Jos\'{e} Carlos Ortiz-Bayliss\orcidID{0000-0003-3408-2166} \and \\
    Hugo Terashima-Mar\'{i}n\orcidID{0000-0002-5320-0773}
}

\authorrunning{J.~A.~Portillo-Quintero et al.}
\titlerunning{A Straightforward Framework For Video Retrieval Using CLIP}

\institute{School of Engineering and Sciences, Tecnologico de Monterrey\\ Av. Eugenio Garza Sada 2501, Monterrey, NL 64849, Mexico\\ a00226024@itesm.mx,\\ \{jcobayliss, terashima\}@tec.mx}

\maketitle

\begin{abstract}
    Video Retrieval is a challenging task where a text query is matched to a video or vice versa. Most of the existing approaches for addressing such a problem rely on annotations made by the users. Although simple, this approach is not always feasible in practice. In this work, we explore the application of the language-image model, CLIP, to obtain video representations without the need for said annotations. This model was explicitly trained to learn a common space where images and text can be compared. Using various techniques described in this document, we extended its application to videos, obtaining state-of-the-art results on the MSR-VTT and MSVD benchmarks\footnote{Code available at: \url{https://github.com/Deferf/CLIP_Video_Representation}}.
\end{abstract}

\section{Introduction}
\label{sec:Introduction}

Video is one of the most consumed forms of media available on the internet. The high consumption of this type of media requires to find suitable methods for finding videos that contain one or more features desired by the users. Most video browsers rely on annotations made by users to identify video contents. Although this solution is simple to implement, it comes at a high price. Relying on annotations to perform a query on videos requires an extensive description of the videos' innards and context. Unfortunately, this information may not be made available. Thus, it is clear that a video retrieval system that can handle user's queries without the need for such annotations represents a relevant topic of study.

This document describes a video retrieval model, which, as its name implies, can retrieve the videos from a collection that are best described by a particular query (text). For example, ``A woman is running" should return videos that contain women running. Given that the Video Retrieval architecture estimates the similarity between video and text, it can also be used to perform the video-to-text retrieval~(VTR) task. It consists of returning captions that best describe the query (video) from a set of description candidates. In either task, the system goal is that, given a query and a set of video-text pairs, it must return the ranking at which the corresponding opposite modality is positioned.

The TVR and VTR tasks can be seen as a method by which video and text contents are funnelled into a fixed-length representation using an embedding function. Since both projections fall in the same dimensional space, a similarity score can be applied, which consequently can be used to rank elements from a set of prospects. Given that similarity metrics between text-video and video-text are equal, TVR and VTR are considered inverse operations. They only depend on the modality of the input prompt.

Some works extensively focus on the video representation by adding pre-trained models considered ``experts''.  Each ``expert" focuses on specific video contents such as sound, face detection, motion, among others. The information from all the experts is multiplexed by a complex gating mechanism~\cite{Gabeur2020MMT,Liu2020CE}. Instead of starting from an elaborated video representation to train a common visual-text space, we propose to use a learned visual-text space to build a video representation. Similarly to Mithun et al.~\cite{Mithun2018MultimodalCuesVSE}, our approach consists of using pre-trained models that measure the similarity between image and text. Then, we extend this idea to handle videos. We experimented with several aggregation methods to comply with the extra temporal dimension.

In this work, we chose CLIP as the base image-text model. CLIP is a state-of-the-art Neural Network, which is pre-trained for image-text pairs~\cite{Radford2021CLIP}. CLIP has proved that similarity learning can be used to train a visual encoder for downstream tasks such as classification, captioning, and clustering, to mention some. We harness the power of its visual representations to create a video representation that can be used directly with its original text encoder to bootstrap a Neural Network model for Video Retrieval. Since our work focuses on aggregation strategies of image features, our method is tested with Zero Shots of the evaluation dataset. Hence, no parameter finetuning is exercised to improve retrieval results.

The remainder of this document is organized as follows. In Section~\ref{sec:Background} we provide the foundations of this investigation and an overview of the most relevant related works. Section~\ref{sec:Experiments} describes the experiments conducted, their main results and their discussion. Finally, in Section~\ref{sec:Conclusion} we present the conclusion and some ideas that may be worth exploring as part of the future work.


\section{Background and Related Work}
\label{sec:Background}

The work presented in this document is related to strategies used to construct a video encoder for video retrieval. It is straight forward to think that image features can serve as a proxy for video representations. In fact, Karpathy et al.~\cite{karpathy2014large} observed that a Convolutional Neural Network~(CNN) feature from a single frame could be discriminative enough for video classification, achieving just 1.3 fewer percentage points than the top accuracy model from the same work, which on its part included more visual and temporal information.

Mithun et al.~\cite{Mithun2018MultimodalCuesVSE} proved that it was possible to supersede the state-of-the-art video retrieval model by obtaining the average visual features obtained from an image-text model. This practice has been implemented on novel models, along with more elaborated video representations. For instance, the state-of-the-art in video retrieval has been pushed by models that implement a Mixture-of-Experts~(MoE) paradigm~\cite{Gabeur2020MMT,Liu2020CE,Miech18MoE,Patrick2021supportset}. The MoE approach proposes a complex video representation by multiplexing the outputs of several pre-trained models (known as ``experts") that attend to particular aspects of video such as motion, face detection, character recognition, among others. 

In this regard, we are aware that at most seven experts have been included in a Video Retrieval model~\cite{Gabeur2020MMT}. Nonetheless, the current state-of-the-art implements a mixture of two experts, indicating that video-text representations may rescind the added complexity that multiple experts convey~\cite{Patrick2021supportset}. Patrick et al.~\cite{Patrick2021supportset} propose that Contrastive Training used by most video retrieval systems encourages repulsive forces on independent, but similar, examples. To alleviate this, they use a support set containing positive examples for each data point on a training batch, so the common video-text space must learn concept sharing. Nonetheless, contrastive training has been proved successful in image and video representation learning~\cite{Chen2020SimCLR,Miech2020end}. 

Contrastive training is a regime on which a model is inducted to pull similar data points together and push apart dissimilar ones on a latent space. The foundational mechanism of the Contrastive Language-Image Pretraining~(CLIP) is the model used in this work. As its name states, the model is pre-trained on 400,000 image-text pairs collected from the Internet. As a siamese neural network, it is composed of an image (ViT-B/32) and text encoder (transformer) that funnel information into a common space where objects can be compared using cosine similarity~\cite{Radford2021CLIP}.

\section{Experiment and Results}
\label{sec:Experiments}

This section describes a mathematical description of CLIP and how we can use it for VTR or TVR. Then, we describe the datasets and metrics considered for this work. Then, we detail the experiments and their main results, followed by a brief discussion of the most relevant findings.

\subsection{CLIP as Video Representation}
\label{subsec:Problem_Formulation}

By using CLIP we obtain the pre-trained functions $\omega{\left(u\right)}=\mathbf{w}$ and $\phi{\left(t\right)}=\mathbf{c_t}$, which encode image $u$ and text $t$ into $\mathbf{w}, \mathbf{c_t} \in \mathbb{R}^d$, where $d = 512$. Assume a video $v$ is composed of $s$ sampled frames such that $v = \left\{ u_{1} , u_{2}, \dots, u_{s} \right\}$. Consequently, we can calculate the embedding of each frame into a matrix $\mathbf{W} \in \mathbb{R}^{d \times s}$ so $\mathbf{W} = \left[ \omega{\left(u_1\right)} = \mathbf{w_{1}}, \mathbf{w_{2}}, \dots, \mathbf{w_{s}} \right] $. Therefore, the problem we try to solve is to find an aggregation function $\Lambda$ that maps the input  $\mathbf{W} \in \mathbb{R}^{d \times s}$ into a video representation $ \mathbf{c_v} \in \mathbb{R}^{d}$. Then, with a video and text representations $\mathbf{c_{v}}$ and $\mathbf{c_{t}}$, we can compute a cosine similarity function (Equation~\ref{eq:CosSimilarity}), which is useful for ranking the video-text pairs inside a dataset given a query of a specific modality.

\begin{equation}
    sim(\mathbf{a},\mathbf{b}) = \frac{\mathbf{a}^T \mathbf{b}}{\|\mathbf{a}\|\|\mathbf{b}\|} 
\label{eq:CosSimilarity}
\end{equation}

\subsection{Datasets}
\label{sec:Datasets}

The proposed framework assumes a set $\mathcal{C}$ of  videos and corresponding captions  pairs in the form $\{\{(v_{i}, t_{ij})\}^n_{i=1}\}^{m(v_i)}_{j=1}$ where the number of captions per video may be non-uniform, hence $m$ is a function of $v$. By design, some datasets are split into sections used for training and validation of results. For the preliminary experiments, we use the training splits to prove our hypothesis, but final results are reported on tests split of their respective datasets.

The datasets involved in this work are listed below.

\begin{description}
    \item [MSR-VTT] is a dataset composed of 10,000 videos, each with a length that ranges from ten to 32 seconds and 200,000 captions. The training, validation and test splits are composed of 6,513, 497 and 2,990 videos, respectively, with 20 corresponding descriptions each~\cite{Xu2016MSR}. The test set has been used in different ways in the literature. Then, we will refer to two common variations as Full~\cite{Liu2020CE} (containing all the 2,990 videos in the test set from MSR-VTT) and 1k-A~\cite{Yu2018JSFusion} (containing only 1000 videos from the 2,990 in the test set in MSR-VTT).
    \item [MSVD] contains 1970 videos, each with a length that ranges from one to 62 seconds. Train, validation and test splits contain 1200, 100 and 670 videos, respectively~\cite{Chen2011MSVD}. Each video has approximately 40 associated sentences in English.
    \item [LSMDC] is comprised 118,081 videos, each with a length that ranges from two to 30 seconds. The videos were extracted from 202 movies. The validation set contains 7,408 videos, and the test set 1,000 videos from movies independent from the training and validation splits~\cite{Rohrbach2015LSMDC}.
\end{description}

All the frames were sampled from each video from the previously mentioned datasets to extract the frame features. Other datasets are related to this work but cannot be used include WIT (WebImageText)~\cite{Radford2021CLIP} and HT100M~\cite{Miech2019HTM100}. WIT is composed of 400,000 image-text pairs on which CLIP was trained on. Since WIT is an image-text dataset that cannot be used as a benchmark for video retrieval. HT100M is a dataset of 100 million video-text pairs, used only as a pre-training set for other Video Retrieval works~\cite{Gabeur2020MMT,Miech2019HTM100,Patrick2021supportset,Rouditchenko2020avlnet}.

\subsection{Metrics}
\label{sec:Metrics}

To conduct our experiments, we follow the testing methodologies used in previous works~\cite{Gabeur2020MMT,Liu2020CE} and report standard retrieval metrics. For median rank (MdR), mean rank (MnR) and standard deviation of rank (StdR), the lower the value, the better the performance. In the case of recall at rank ($R@k$, where $k = \{1, 5, 10\}$), the higher the value, the better the performance. For datasets that involve multiple sentences per video ---such as Full from MSR-VTT and MSVD test---, we follow the protocol used by Liu et al.~\cite{Liu2020CE} and use the minimum rank among all associated sentences to a given video query.

\subsection{Exploratory Experiments}
\label{sec:ExploratoryExperiments}

In the exploratory experiments, we empirically define two candidates for frame-level aggregation $\Lambda$ functions. We conduct this set of preliminary experiments on a validation sample comprised of 1,000 video-text pairs from MSR-VTT. The first frame-level aggregation function is based on the idea that it is feasible to obtain reasonable video representations by only considering one frame sample~\cite{karpathy2014large}. Given the feature matrix $\mathbf{W} \in \mathbb{R}^{d \times s}$, we define $\Lambda_{s}(W) = W_{30} \in \mathbb{R}^{d}$ as a function that returns the features of the 30\textsuperscript{th} frame. Since these videos contain approximately 30 frames per second, this is equivalent to sampling a frame from the first second of the video. 

A second candidate for an aggregation function is proposed by Mithun et al.~\cite{Mithun2018MultimodalCuesVSE}, who suggest that the average of frame-level features from videos can be used as an approximator for video representations. This method has extensively been used in other retrieval works~\cite{Gabeur2020MMT,Liu2020CE,Miech2020end,Miech2019HTM100,Patrick2021supportset}. Consequently, we define $\Lambda_{avg}(\mathbf{W}) = \Bar{W} \in \mathbb{R}^{d}$, where $\Bar{W}$ is the average value of matrix columns.

Given that videos present dynamic events in which several sequences of frames can represent completely different things, we used $k$-means as the method for aggregation~\cite{sun2019videobert}. With this implementation, the aggregation function follows the form $\Lambda_{k}(\mathbf{W}) = W \in \mathbb{R}^{d \times k}$, which returns $k$ video embeddings. For evaluation purposes, we repeat the ranking procedure with the obtained independent video representations $k$ times and register each query's minimum rank, then calculate the retrieval metrics. 

\begin{table}[ht!]
\centering
\begin{tabular}{ccccccc}
\toprule
\textbf{$\Lambda$} & \textbf{R@1} & \textbf{R@5} & \textbf{R@10} & \textbf{MdR} & \textbf{MnR} & \textbf{StdR} \\ 
\midrule
$\Lambda_{s}$ & 24.9 & 46.1 & 56.9 & 7.0 & 64.61 & 149.21 \\
$\Lambda_{avg}$ & 35.4 & 58.0 & 67.2 & 3.0 & 39.81 & 111.43 \\
$\Lambda_{2}$ & 34.3 & 57.8 & 66.5 & 3.0 & 40.23 & 112.85 \\
$\Lambda_{3}$ & 34.4 & 57.7 & 66.6 & 3.0 & 39.77 & 110.69 \\
$\Lambda_{4}$ & 33.7 & 58.4 & 66.9 & 3.0 & 37.98 & 107.53 \\
$\Lambda_{5}$ & 34.4 & 57.6 & 66.1 & 3.0 & 38.44 & 108.02 \\
$\Lambda_{6}$ & 34.9 & 58.4 & 67.6 & 3.5 & 37.44 & 108.34 \\
$\Lambda_{7}$ & 35.3 & 58.1 & 67.5 & 4.0 & 38.33 & 107.88 \\
$\Lambda_{8}$ & 33.9 & 57.7 & 67.9 & 3.0 & 38.23 & 107.32 \\
$\Lambda_{9}$ & 33.9 & 57.2 & 67.1 & 3.0 & 37.87 & 108.23 \\
$\Lambda_{10}$ & 35.0 & 57.8 & 68.0 & 3.0 & 37.26 & 107.34 \\
\bottomrule
\end{tabular}
\caption{Text-to-Video Retrieval results on the MSR-VTT validation set, using different aggregation functions.}
\label{table:MSRLambda}
\end{table}

Based on the results depicted in Table~\ref{table:MSRLambda}, the average-based methods obtain the best results in terms of the metrics used. It is noticeable that, among $k$-means methods, there is no significant difference between the results. This may be because MSR videos do not exceed 32 seconds in length, which may not be enough to differentiate the centroids when creating the clusters. We appeal to Occam's Razor principle regarding the aggregation method and select $\Lambda_{avg}$ for further experiments since it accomplishes a similar performance to $k$-means based aggregation methods but with a lower calculation complexity.

\subsection{Confirmatory Experiments}
\label{sec:ConfirmatoryExperiments}

This section compares our video retrieval model against the state-of-the-art in the MSR-VTT, MSVD and LSMDC datasets. In all the cases, we evaluate both the TVR and VTR tasks.

In MSR-VTT,  we are able to supersede the R@1 score of the previous best model SSB~\cite{Patrick2021supportset} on the split 1k-A for the TVR task. However, we are positioned behind previous works on other recall metrics (Table~\ref{table:BigMSRVTT}). Besides, we consistently achieve state-of-the-art results on all the recall metrics in the Full split from MSR-VTT. In the MSVD dataset, we obtain state-of-the-art results on most of the retrieval metrics (Table \ref{table:BigMSVD}). We suppose that models that are based on MoE such as SSB~\cite{Patrick2021supportset} and CE~\cite{Liu2020CE} cannot use all of their implemented experts because the videos in MSVD lack audio information, so they have to rely only on visual features. In LSMDC, we do not obtain state-of-the-art results, but we are positioned second-best (Table \ref{table:BigLSMDC}). Given that video descriptions in this dataset do not follow the form of a typical sentence, as they are designed to teach a model to recognize characters and interactions between movie scenes, we commend the robustness of CLIP's text encoder because it could adapt to a new sentence schema.

\begin{table}[ht!]
\centering
\begin{tabular}{@{}lcccccccccc@{}}
\cmidrule(l){4-11}
\textbf{} & \multicolumn{1}{l}{\textbf{}} &  & \multicolumn{4}{c}{TVR} & \multicolumn{4}{c}{VTR} \\ \midrule
\multicolumn{1}{c}{Method} & Training & Test Set & \textbf{\textbf{R@1}} & \textbf{\textbf{R@5}} & \textbf{\textbf{R@10}} & \textbf{\textbf{MdR}} & \textbf{R@1} & \textbf{R@5} & \textbf{R@10} & \textbf{MdR} \\ \midrule
JSFusion~\cite{Yu2018JSFusion} & M & 1k-A & 10.2 & 31.2 & 43.2 & 13 & - & - & - & - \\
HT100M~\cite{Miech2019HTM100} & H+M & 1k-A & 14.9 & 40.2 & 52.8 & 9 & 16.8 & 41.7 & 55.1 & 8 \\
CE~\cite{Liu2020CE} & M & 1k-A & 20.9 & 48.8 & 62.4 & 6 & 20.6 & 50.3 & 64 & 5.3 \\
AVLnet~\cite{Rouditchenko2020avlnet} & H+M & 1k-A & 27.1 & 55.6 & 66.6 & 4 & 28.5 & 54.6 & 65.2 & 4 \\
MMT~\cite{Gabeur2020MMT} & H+M & 1k-A & 26.6 & 57.1 & \textbf{69.6} & 4 & 27.0 & 57.5 & 69.7 & 3.7 \\
SSB~\cite{Patrick2021supportset} & H+M & 1k-A & 30.1 & \textbf{58.5} & 69.3 & \textbf{3} & \textbf{28.5} & \textbf{58.6} & \textbf{71.6} & \textbf{3} \\
CLIP & W & 1k-A & \textbf{31.2} & 53.7 & 64.2 & 4 & 27.2 & 51.7 & 62.6 & 5 \\ \midrule
VSE~\cite{Mithun2018MultimodalCuesVSE} & M & Full & 5.0 & 16.4 & 24.6 & 47 & 7.7 & 20.3 & 31.2 & 28 \\
VSE++~\cite{Mithun2018MultimodalCuesVSE} & M & Full & 5.7 & 17.1 & 24.8 & 65 & 10.2 & 25.4 & 35.1 & 25 \\
Multi Cues~\cite{Mithun2018MultimodalCuesVSE} & M & Full & 7.0 & 20.9 & 29.7 & 38 & 12.50 & 32.10 & 42.4 & 16 \\
W2VV~\cite{Dong2018W2VV} & M & Full & 6.1 & 18.7 & 27.5 & 45 & 11.8 & 28.9 & 39.1 & 21 \\
Dual Enc.~\cite{Dong2019dual} & M & Full & 7.7 & 22.0 & 31.8 & 32 & 13.0 & 30.8 & 43.3 & 15 \\
E2E~\cite{Miech2020end} & M & Full & 9.9 & 24.0 & 32.4 & 29.5 & \textbf{-} & \textbf{-} & \textbf{-} & \textbf{-} \\
CE~\cite{Liu2020CE} & M & Full & 10.0 & 29.0 & 42.2 & 16 & 15.6 & 40.9 & 55.2 & 8.3 \\
CLIP & W & Full & \textbf{21.4} & \textbf{41.1} & \textbf{50.4} & \textbf{10} & \textbf{40.3} & \textbf{69.7} & \textbf{79.2} & \textbf{2} \\ \bottomrule
\end{tabular}
\caption{TVR and VTR results in the MSR-VTT dataset. M, H and W denote training on MSR-VTT, HT100M and WIT, respectively.}
\label{table:BigMSRVTT}
\end{table}

\begin{table}[ht!]
\centering
\begin{tabular}{lccccccccc}
\cmidrule(l){3-10}
                                                      & \multicolumn{1}{l}{} & \multicolumn{4}{c}{\textbf{TVR}}                   & \multicolumn{4}{c}{\textbf{VTR}}                   \\ \hline
\multicolumn{1}{c}{\textbf{Method}}                   & \textbf{Training}    & \textbf{R@1}  & \textbf{R@5}  & \textbf{R@10} & \textbf{MdR} & \textbf{R@1}  & \textbf{R@5}  & \textbf{R@10} & \textbf{MdR} \\ \hline
VSE   \cite{Mithun2018MultimodalCuesVSE}              & D                    & 12.3          & 30.1          & 42.3          & 14           & 34.7          & 59.9          & 70.0          & 3            \\
VSE++ \cite{Mithun2018MultimodalCuesVSE}              & D                    & 15.4          & 39.6          & 53.0          & 9            & -             & -             & -             & -            \\
Multi Cues   \cite{Mithun2018MultimodalCuesVSE}       & D                    & 20.3          & 47.8          & 61.1          & 6            & -             & -             & -             & -            \\
CE \cite{Liu2020CE}                                   & D                    & 19.8          & 49.0          & 63.8          & 6            & -             & -             & -             & -            \\
Support-set Bottleneck   \cite{Patrick2021supportset} & H+D                  & 28.4          & 60.0          & 72.9 & 4            & -             & -             & -             & -            \\ \hline
CLIP                                                  & W                    & \textbf{37} & \textbf{64.1} & \textbf{73.8}          & \textbf{3}   & \textbf{59.9} & \textbf{85.2} & \textbf{90.7} & \textbf{1}   \\ \hline
\end{tabular}
\caption{TVR and VTR results in the MSVD dataset. D, H and W denote training on MSVD, HT100M and WIT, respectively.}
\label{table:BigMSVD}
\end{table}

\begin{table}[ht!]
\centering
\begin{tabular}{lccccccccc}
\cmidrule(l){3-10}
\multicolumn{1}{l}{}           & \multicolumn{1}{l}{} & \multicolumn{4}{c}{\textbf{TVR}}                    & \multicolumn{4}{c}{\textbf{VTR}}                   \\ \hline
\textbf{Method}                & \textbf{Training}    & \textbf{R@1}  & \textbf{R@5}  & \textbf{R@10} & \textbf{MdR}  & \textbf{R@1}  & \textbf{R@5}  & \textbf{R@10} & \textbf{MdR} \\ \hline
JSFusion \cite{Yu2018JSFusion} & L                    & 9.1           & 21.2          & 34.1          & 36            & \textbf{12.3} & \textbf{28.6} & \textbf{38.9} & \textbf{20}  \\
CE \cite{Liu2020CE}            & L                    & 11.2          & 26.9          & 34.8          & 25.3          & -             & -             & -             & -            \\
MMT \cite{Gabeur2020MMT}       & H+L                  & \textbf{12.9} & \textbf{29.9} & \textbf{40.1} & \textbf{19.3} & -             & -             & -             & -            \\ \hline
CLIP                           & W                    & 11.3          & 22.7          & 29.2          & 56.5          & 6.8           & 16.4          & 22.1          & 73           \\ \hline
\end{tabular}
\caption{TVR and VTR results in the LSMDC dataset. L, H and W denote training on LSMDC, HT100M and WIT, respectively.}
\label{table:BigLSMDC}
\end{table}

\subsection{Discussion}
\label{sec:Discussion}

\begin{figure}[ht!]
    \centering
    \includegraphics[height=2in]{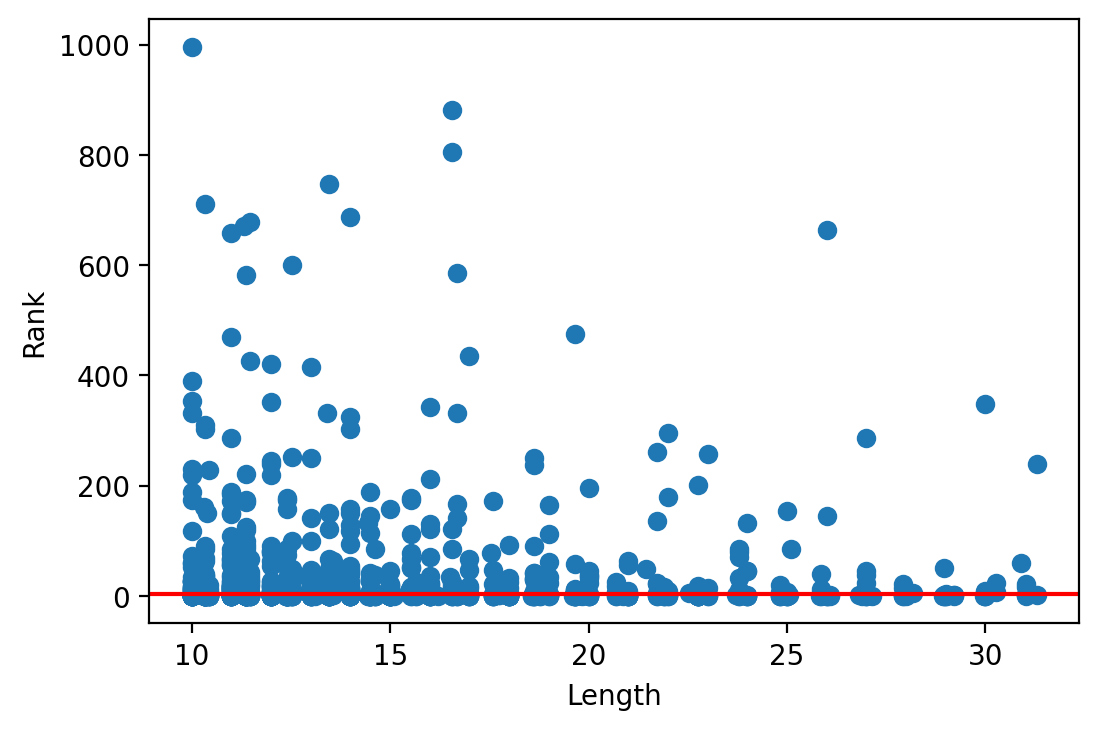}
    \caption{Scatter plot of video length and assigned rank on TVR task on the 1k-A test split. The red line represents the median rank.}
    \label{fig:ScatterPlot}
\end{figure}

Although we obtain outstanding results on different metrics and datasets, there are some things worth discussing. For example, our original supposition was that the ranking worsens as the video gets longer. To confirm or reject this idea, we produced Figure~\ref{fig:ScatterPlot}. Figure~\ref{fig:ScatterPlot} depicts the video length in seconds ($x$-axis), and the rank assigned to it ($y$-axis). As a video gets longer, we expected that it would be more difficult for the video representation to capture the temporal elements. Hence it would be ranked worse. However, the experiment conducted on set 1k-A from MSR-VTT shows that ranking varies wildly independently from video length (at least as long as the videos present in the dataset).

We proceeded to look at the worst ranked video-text pairs, we noticed that several sentences incorporated phrases like ``a family is having a conversation" or ``a man talking about a woman" hinting that sentences that were mainly describing audio content would be ranked worse. This conclusion is reinforced by the fact that our model scored the best on MSVD, a dataset that by design does not contain any audio track and text descriptions are based on what can be visualized.

\section{Conclusion and Future Work}
\label{sec:Conclusion}

This work presents the first implementation of CLIP to obtain video features. Our method works by leveraging its learned common image-text space without the need for parameter finetuning (Zero-Shot). We apply an aggregation function to frame-level features, common in other video retrieval works. Our work focuses only on visual and text modalities,
as it supersedes methods that implement a complex mixture of pre-trained models obtaining state-of-the-art results on the MSVD and MSR-VTT datasets. 

One potential application of this CLIP-derived implementation is to retrieve specific moments inside videos. Also, it is yet unseen how will our video representation behave if tested as a video classifier. This methodology might be useful to create a video representation that is based on CLIP for longer durations. For example, other works have used frame features to construct a graph that can change through time~\cite{mao2018hierarchical}. Such a representation could keep the strong text alignment suitable for video retrieval. Also, our work can be used as an expert on a future MoE video retrieval system.

\section*{Acknowledgments}

This research was partially supported by ITESM Research Group with Strategic Focus on Intelligent Systems.

\bibliographystyle{splncs04}
\bibliography{references}

\end{document}